\documentclass[11pt,a4paper]{article}
\usepackage{lettrine}
\usepackage[dvipsnames]{xcolor}
\usepackage[nohyperref]{emnlp2017}
\usepackage{times}
\usepackage{latexsym}
\usepackage{multirow}
\usepackage{paralist}
\usepackage{url}

\usepackage[utf8]{inputenc}
\usepackage[T1]{fontenc}
\usepackage{multirow}

\usepackage{graphicx}               
\usepackage{amssymb}
\usepackage{amsthm}
\usepackage{mathtools}
\usepackage{enumerate}
\usepackage[hang]{footmisc}
\usepackage{tablefootnote}
\usepackage{marvosym}
\usepackage{xspace}
\usepackage{calc}

\usepackage[british]{babel}

\emnlpfinalcopy

\def\emnlppaperid{***}

\newcommand\tofrom{$\leftrightarrow$\ }
\newcommand\bleu{{\sc Bleu}\xspace}

\title{The University of Edinburgh's Neural MT Systems for WMT17}

\author{Rico Sennrich,  Alexandra Birch, Anna Currey, Ulrich Germann, \\
\bf Barry Haddow, Kenneth Heafield,
 Antonio Valerio Miceli Barone \and  Philip Williams \\
University of Edinburgh, Scotland \\
  }

\date{}

\begin{document}

\maketitle

\begin{abstract}
This paper describes the University of Edinburgh's submissions to the WMT17 shared news translation and biomedical translation tasks.
We participated in 12 translation directions for news, translating between English and Czech, German, Latvian, Russian, Turkish and Chinese.
For the biomedical task we submitted systems for English to Czech, German, Polish and Romanian.
Our systems are neural machine translation systems trained with Nematus, an attentional encoder-decoder.
We follow our setup from last year and build BPE-based models with parallel and back-translated monolingual training data.
Novelties this year include the use of deep architectures, layer normalization, and more compact models due to weight tying and  improvements in BPE segmentations.
We perform extensive ablative experiments, reporting on the effectivenes of layer normalization, deep architectures, and different ensembling techniques.
\end{abstract}

\section{Introduction}

We participated in the WMT17 shared news translation task for 12 translation directions, translating between English and Czech, German, Latvian, Russian, Turkish and Chinese, and in the 
WMT17 shared biomedical translation task for English to Czech, German, Polish and Romanian.\footnote{We provide trained models and training commands at\\ \url{http://data.statmt.org/wmt17_systems/}}
We submitted neural machine translation systems trained with Nematus \citep{nematus}.
Our setup is based on techniques described in last year's system description \citep{sennrich-haddow-birch:2016:WMT}, including the use of subword models \citep{DBLP:journals/corr/SennrichHB15}, back-translated monolingual data, \citep{2015arXiv151106709S}, and re-ranking with right-to-left models.

This year, we experimented with deep network architectures, new ways to include monolingual data, and different ensembling variants.
Other novelties include obtaining more compact models via better BPE segmentation and by weight tying \cite{DBLP:journals/corr/PressW16},
and speeding up model training with layer normalization \citep{DBLP:journals/corr/BaKH16} and adam \citep{DBLP:journals/corr/KingmaB14}.

We perform extensive ablative experiments across language pairs to evaluate the effectiveness of each of these approaches.
When comparing this year's baseline models to our best results, we show consistent increases in scores of
 2.2--5 {\sc Bleu}  for our 12 news task language pairs. Among the constrained submissions to the news task, our submissions are ranked tied first for 11 out of the 12 translation directions in which we participated. For the biomedical task, we obtained the highest {\sc Bleu} for all our submitted systems.

For the 6 language pairs for which we participated both in WMT16 and WMT17, we also show the scores of last year's systems.
We observe solid improvements with increases of 1.5--3 {\sc Bleu} for single models.
Some of these improvements are due to differences in training data, preprocessing and hyperparameters, but most of the 
increase is due to layer normalization and deeper models. It is worth mentioning that our deeper models were trained on single GPUs, showing that the benefits of deeper models can be harnessed with limited hardware resources.

\section{Novelties}

Here we describe the main differences to last year's systems.

\subsection{Subword Segmentation}
\label{bpe-section}

Like last year, we use joint byte-pair encoding (BPE) for subword segmentation \citep{DBLP:journals/corr/SennrichHB15} (except for ZH$\leftrightarrow$EN, where we train two separate BPE models).
Joint BPE introduces undesirable edge cases in that it may produce subword units that have only been observed in one side of the parallel training corpus, and may thus be out-of-vocabulary at test time.
To prevent this, we have modified our BPE script to only produce subword units at test time that have been observed in the source side of the training corpus.\footnote{\scriptsize \url{https://github.com/rsennrich/subword-nmt}}
Out-of-vocabulary subword units are recursively segmented into smaller units until this condition is met.

We use the same technique to disallow rare subword units (words occurring less than 50 times in the training corpus), both at test time and in the training corpus, both on the source-side and the target-side.
This reduces the number of vocabulary symbols reserved for spurious, low-frequency subword units, and allows for more compact models.
For example, for  EN$\leftrightarrow$DE, using 90000 joint BPE operations, this filtering reduces the network vocabulary size for English from 80581 to 51092, with only a minor increase in sequence length (+0.2\%).
In preliminary experiments, this did not significantly affect \bleu, but slightly reduced the number of spurious OOVs produced -- on EN$\rightarrow$DE, unigram precision for OOVs increased from 0.34 to 0.36 on newstest2015 ($N=1168$).

\subsection{Layer Normalisation and Adam}

This year, we use layer normalisation \citep{DBLP:journals/corr/BaKH16} for all systems.
We apply layer normalisation to all recurrent and feed-forward layers, except for layers that are followed by a softmax.
As SGD optimization algorithm, we use adam \citep{DBLP:journals/corr/KingmaB14} instead of adadelta \citep{DBLP:journals/corr/abs-1212-5701}, which we used last year.

In preliminary experiments, we found that both adam and layer normalisation lead to faster convergence, and result in better performance.

\subsection{Deep Architectures}

\citet{micelibarone2017} describe different deep recurrent architectures for neural machine translation.
We use some of these architectures for our shared task submissions.
We mainly use a \emph{deep transition} architecture, but some runs use a \emph{stacked} architecture.
Implementations of both of these architectures are available in Nematus.

For completeness, we here reproduce the description of the relevant deep architectures from \cite{micelibarone2017}.
Note that some results reported by \citet{micelibarone2017} were obtained after the shared task submission, which explains why we did not choose the best-performing architecture, BiDeep.

\subsubsection{Deep Transition Architecture}

As in a baseline shallow Nematus system, the encoder is a bidirectional recurrent neural network.
However, instead of being a simple GRU transition \citep{DBLP:journals/corr/ChoMGBSB14}, the recurrence transition is itself composed of multiple GRU transitions with independently trainable parameters, all of which are executed sequentially for each input word.

Let $L_s$ be the encoder recurrence depth, then for the $i$-th source word in the forward direction the forward source word state $\overrightarrow{h}_i \equiv \overrightarrow{h}_{i,L_s}$ is computed as
\begin{align*}
\overrightarrow{h}_{i,1} &= \text{GRU}_1   \left( x_{i}, \overrightarrow{h}_{i-1,L_s}  \right) \\
\overrightarrow{h}_{i,k} &= \text{GRU}_k   \left(0, \overrightarrow{h}_{i,k-1}  \right) \text{for } 1 < k \leq L_s
\end{align*}
where the input to the first GRU transition is the word embedding $x_{i}$, while the other GRU transitions have no external inputs.
Note that each GRU transition is not internally recurrent, recurrence only occurs at the level of the whole multi-layer transition cell, as the previous word state $\overrightarrow{h}_{i-1,L_s}$ enters the computation in the first GRU transition. \\
The reverse source word states are computed similarly and concatenated to the forward ones to form the bidirectional source word states $C \equiv \left\lbrace \left[\overrightarrow{h}_{i,L_s} \overleftarrow{h}_{i,L_s} \right] \right\rbrace$.

The deep transition decoder is obtained by extending the baseline decoder in a similar way.
Note that the baseline decoder of Nematus has already a transition depth of two, with the first GRU transition receiving as input the embedding of the previous target word and the second GRU transition receiving as input a context vector computed by the attention mechanism.
We extend this decoder architecture to an arbitrary transition depth $L_t$ as follows
\begin{align*}
s_{j,1} &= \text{GRU}_1   \left( y_{j-1}, s_{j-1,L_t}  \right) \\
s_{j,2} &= \text{GRU}_2   \left( \text{ATT}(C, s_{j,1}), s_{j,1}  \right) \\
s_{j,k} &= \text{GRU}_k   \left(0, s_{j,k-1}  \right) \text{for } 2 < k \leq L_t
\end{align*}
where $y_{j-1}$ is the embedding of the previous target word and $\text{ATT}(C, s_{i,1})$ is the context vector computed by the attention mechanism.
GRU transitions other than the first two do not have external inputs.
The target word state vector $s_j \equiv s_{j, L_t}$ is then used by the feed-forward output network to predict the current target word.

In our experiments we use an encoder recurrence depth $L_s = 4$ and a decoder recurrence depth $L_t = 8$.

\subsubsection{Stacked architecture}
For our stacked architecture we use a variation of the one proposed by \newcite{zhou2016deep} with residual connections between the stack layers.

The forward encoder consists of a stack of GRU recurrent neural networks, the first one processing words in the forward direction, the second one in the backward direction, and so on, in alternating directions.
For an encoder stack depth $D_s$, and a source sentence length $N$, the forward source word state $\overrightarrow{w}_i \equiv \overrightarrow{w}_{i,D_s}$ is computed as
\begin{align*}
\overrightarrow{w}_{i,1} &= \overrightarrow{h}_{i,1} = \text{GRU}_1   \left( x_{i}, \overrightarrow{h}_{i-1,1}  \right) \\
\overrightarrow{h}_{i,2k} &= \text{GRU}_{2k}   \left(\overrightarrow{w}_{i,2k-1}, \overrightarrow{h}_{i+1,2k}  \right) \\
  &\text{ for } 1 < 2k \leq D_s \\
\overrightarrow{h}_{i,2k+1} &= \text{GRU}_{2k+1}   \left(\overrightarrow{w}_{i,2k}, \overrightarrow{h}_{i-1,2k+1}  \right) \\
  &\text{ for } 1 < 2k+1 \leq D_s \\
\overrightarrow{w}_{i,j} &= \overrightarrow{h}_{i,j} + \overrightarrow{w}_{i,j-1} \\
& \text{ for } 1 < j \leq D_s
\end{align*}
where we assume that source word indexes $i$ start at $0$ and $\overrightarrow{h}_{0,k}$ and $\overrightarrow{h}_{N+1,k}$ are zero vectors.
Contrary to the deep transition encoder, each GRU transition here is a recurrent cell by itself.
Note the residual connections: at each level above the first one, the word state of the previous level $\overrightarrow{w}_{i,j-1}$ is added to the recurrent state of the GRU cell $\overrightarrow{h}_{i,j}$ to compute the the word state for the current level $\overrightarrow{w}_{i,j}$.
The backward encoder has the same structure, except that the first layer of the stack processes the words in the backward direction and the subsequent layers alternate directions.
The forward and backward word states are then concatenated to form bidirectional word states $C \equiv \left\lbrace \left[\overrightarrow{w}_{i,D_s} \overleftarrow{w}_{i,D_s} \right] \right\rbrace$.

The stacked decoder has a similar structure, without any direction alternation.
While the base GRU in the decoder is a conditional GRU with two transitions, we use simple GRUs on higher layers.
The "external" input to the higher layers is the concatenation of the state below and the context vector from the base RNN \cite{wu2016google}.

Where we have used the stacked architecture, we set both encoder and decoder depths to 4.

\subsection{Monolingual Data}
\label{sec:monolingual}

Like last year, we use back-translated monolingual data to augment our training data sets.
We use two different training regimes to incorporate this monolingual data. 
In the \emph{mixed} approach, we mix the synthetic and parallel data from the beginning of training, whilst in the \emph{fine-tuned} approach we train
a system using only the parallel data, then when this has converged we continue training using the parallel/synthetic mix. 
In both cases, the mixing proportions are set to 1:1, over-sampling from the smaller corpus where necessary. 
We find that the \emph{mixed} approach is faster overall to train,  although the fine-tuned approach  has the advantage that the intermediate model could be adapted to a different domain using appropriate in-domain data.

For EN$\leftrightarrow$TR, we also experiment with a novel approach for incorporating target-side monolingual data. 
This consists of copying a monolingual corpus to convert it into a bitext where the source and target sides are identical. 
This copied bitext is then mixed with the parallel and back-translated data in order to train the NMT system; no distinciton is made between the copied, back-translated, and parallel data during training. 
The mixing proportions of parallel, copied, and back-translated data we use in the EN$\leftrightarrow$TR experiments are 1:2:2, and we use the same monolingual data for both the copied and the back-translated corpora. 
More details can be found in~\newcite{currey2017}. 

Table~\ref{copied} shows the results of using the copied monolingual data while training. 
All systems are trained using parallel and back-translated data. 
Adding the copied monolingual data either yields modest improvements or does no damage, so we adopt this strategy for all EN$\leftrightarrow$TR experiments.

\begin{table}
\centering
\small
\caption{BLEU scores for EN$\leftrightarrow$TR when adding copied monolingual data.}
\label{copied}\vspace{1em}

\begin{tabular}{l|cc|cc}
& \multicolumn{2}{c|}{\bfseries TR$\to$EN} & \multicolumn{2}{c}{\bfseries EN$\to$TR}\\
\bfseries system & \bfseries 2016 & \bfseries 2017 & \bfseries 2016 & \bfseries 2017\\
\hline
baseline &
20.0 & 19.7 & 
13.2 & 14.7 \\ 
+copied &
20.2 & 19.7 & 
13.8 &  15.6 \\ 
\hline
\end{tabular}
\end{table}

In preliminary experiments, we applied the same approach for EN$\leftrightarrow$LV (in a 1:1:1 ratio).
Compared to our baseline EN$\leftrightarrow$LV systems, the addition of copied monolingual led to a slight decrease in translation quality (around 0.5 BLEU) on the devset and a slight improvement (0.1 BLEU) on the newstest2017 set.

\subsection{Memory Efficiency}

We reduce the memory footprint of our models by reducing the vocabulary sizes (Section \ref{bpe-section}), and by tying the weights of the target-side embedding and the transpose of the output weight matrix,
which have the same dimensionality in our architecture \citep{DBLP:journals/corr/PressW16}.
Using these techniques, we were able to train deep  models on single GPUs (equipped with 8--12GB memory) without requiring model parallelism.

\section{System Overview}

\subsection{Data and Preprocessing}
\label{sec:preprocess}

All our systems are constrained and use data from the website of the shared task.\kern-.25ex\footnote{\url{http://data.statmt.org/wmt17/translation-task.html}}

For preprocessing, we use the Moses tokenizer with hyphen splitting ("-a" option),
and perform truecasing with Moses scripts \citep{koehnmoses}.
For subword segmentation, we use 90000 joint BPE operations, filtered according to section\ \ref{bpe-section}.
The preprocessing pipeline was different for Russian and Chinese (because of non-Latin scripts).
For EN$\rightarrow$LV, we used the data that was prepared for the QT21 system combination \citep{Jan-Thorsten2017}.
The variations are described in the language-specific sections below.

\subsection{Baseline Systems}

We train all systems with Nematus \citep{nematus}, which implements an attentional encoder-decoder with small modifications to the model in \citet{DBLP:journals/corr/BahdanauCB14}.
We use word embedding sizes of 500 or  512, and hidden layer size 1024.
We adapt the size of the network vocabulary to the size of the BPE vocabulary of the respective language.

We use adam \citep{DBLP:journals/corr/KingmaB14} as optimizer with a learning rate of 0.0001, and a batch sizes of 60 or 80 (depending on GPU memory).
We filter out sentences with a length greater than 50 subwords.
We tie the weights of the target-side embedding and the transpose of the output weight matrix \citep{DBLP:journals/corr/PressW16}.
We stop training when the validation cross-entropy fails to reach a new minimum for 10 consecutive save-points (saving every 10000 updates) and select the final model as the one having the best \bleu\ on validation.

For ensembling, we contrast two strategies:

\begin{itemize}
\item {\bfseries checkpoint ensembles}, i.e.\ using the last $N$ checkpoints of a single training run, which is a cheap way of obtaining an ensemble, and which we used in last year's submission.
\item {\bfseries independent ensembles}, i.e.\ training $N$ models independently, potentially with different hyperparameters, which is more expensive, but likely to yield more diversity.
\end{itemize}

\section{Experiments}

\subsection{Chinese \tofrom English}
For this language pair, we use all the available parallel data, except for 2000 sentence pairs from news-commentary which we hold back for validation. The English side is
preprocessed using the same pipeline as for other language pairs, training a single BPE model with 59500 merge operations. For the Chinese side, we segment it
using the Jieba\footnote{\url{https://github.com/fxsjy/jieba}} segmenter, except for Books 1--10 and data2011 which were already segmented.
We then learn a BPE model on the segmented Chinese, also using 59500 merge operations.

For training the ZH$\rightarrow$EN system we augment the parallel data with back-translation of the WMT news2016 monolingual corpus, translated using a shallow Nematus
system built from the parallel data only. Since there was no monolingual news release for Chinese, we use LDC Chinese Gigaword (4th edition) to create synthetic data 
for EN$\rightarrow$ZH, again
using a shallow Nematus system for back-translation.
In total we have approximately 24M parallel sentences, plus 20M sentences of synthetic data for ZH$\rightarrow$EN and 8.5M for EN$\rightarrow$ZH.

For ZH$\rightarrow$EN we run 3 separate training runs in each direction (i.e. target left-right and target right-left). One run in each direction
uses the stacked model architecture, and the fine-tuned training regime, whereas the other two use the deep transition architecture with a mixed training regime. 
Initially we used a 2000 sentence portion of news-commentary for validation, but during the fine-tuning phase of the fine-tuned runs we switch to the development
set released for the task (newsdev2017). 
For the mixed runs, we found that training converged and started to overfit the news-commentary validation set, so we restart with newsdev2017 as
validation and ran to convergence. The final system is an ensemble of the best validation \bleu\ model from each of the three target left-right runs,
rescored with the three target right-left runs, and reranked.

For EN$\rightarrow$ZH we use the same training setup, with three runs for each target direction, and the same mix of models and training regimes.
We use news-commentary as the validation set, except during the fine-tuning phase, where we use newsdev2017. The final system is an ensemble of four target left-right systems
(the best validation \bleu\ model from each of the three runs, plus the same from the first run before fine-tuning), rescored with a similar ensemble
of target right-left models and reranked. The final output is post-processed by removing all spaces (except when there was an ascii letter on either
side) and then converting ascii full-stops and commas to their appropriate CJK unicode equivalents.

\subsection{Czech \tofrom English}
To create the parallel corpus, we take the whole of CzEng 1.6pre \cite{czeng16:2016}, plus the latest WMT releases of Europarl, News-commentary and CommonCrawl. We clean
the corpus by running langid\footnote{\url{https://github.com/saffsd/langid.py}} over both sides and rejecting any parallel sentences whose English sides
are not labelled as English, or whose Czech sides are not labelled as Czech, Slovak or Slovenian, by langid.\footnote{Since langid does not use an estimate of prior
language probability, this is a crude way of improving recall.}  The rest of the preprocessing pipeline is the same as the general case (Section \ref{sec:preprocess}).

The parallel training data is augmented with synthetic parallel data created from the WMT news2016 monolingual corpus, back-translated using Edinburgh's WMT16 systems.\kern-.25ex\footnote{The binary models are available at \url{http://data.statmt.org/rsennrich/wmt16_systems/}}
This provides about 20M synthetic parallel sentences for CS$\rightarrow$EN and nearly 6M for EN$\rightarrow$CS.

For CS$\rightarrow$EN we use the stacked model architecture for all systems, training 4 target left-right systems and 4 target right-left.
The first of the 
left-right systems use the fine-tuned training regime, whereas the rest are all trained using the mixed regime. The final system is an ensemble of
the left-right systems, with 12-best lists rescored with the right-left systems and reranked.

The same number of left-right and right-left models are used for EN$\rightarrow$CS, with 2 of the left-right models using the fine-tuned training regime
and the rest the mixed training regime. One of these fine-tuned models uses the stacked architecture whilst all the other EN$\rightarrow$CS models use
the deep transition architecture. Once again, the final system is an ensemble of the left-right models, rescored and reranked with the right-left systems.

\subsection{German \tofrom English}
For the German-to-English and English-to-German system we use the
pre-processed training data sets for the shared news translation task
provided by the task
organizers,\kern-.25ex\footnote{\url{http://data.statmt.org/wmt17/translation-task/preprocessed/de-en/}}
and supplement them with synthetic training data
\citep{2015arXiv151106709S} created by back-translating
ca. 10 million sentences each from the 2016 monolingual news
crawl data sets available through the web site for the shared
task.
For back-translation, we use Edinburgh's WMT16 systems.

Eight independent deep models are trained for each translation
direction: four producing the translation left-to-right; four
producing it right-to-left. The left-to-right models are ensembled to
produce an n-best list of 50 translation hypotheses (beam size 50),
which are then re-ranked by an ensemble of the four right-to-left
models.

\subsection{Latvian \tofrom English}
For EN$\rightarrow$LV, we use the parallel data that was prepared for the QT21 system combination \citep{Jan-Thorsten2017}.
The main differences from the standard preprocessing pipeline described in Section~\ref{sec:preprocess} are the use of a custom tokenizer with Latvian-specific handling of abbreviations, dates, numeric expressions, etc. and data filtering to remove noisy sentence pairs.
For LV$\rightarrow$EN, we apply  the standard preprocessing pipeline after filtering the parallel corpus to remove the noisy sentence pairs identified during EN$\rightarrow$LV preparation.
In both cases, we reserve the first 2000 sentences of the (unfiltered) LETA news corpus for use as a validation set during system development (with newsdev2017 used as a test set).

To produce LV$\rightarrow$EN and EN$\rightarrow$LV synthetic data we back-translate the WMT monolingual English and Latvian news 2016 corpora, respectively.
We use phrase-based systems for back-translation, since these produced better translations (according to BLEU) than our preliminary parallel-only neural systems.
The EN$\rightarrow$LV synthetic data was subsequently filtered using the same method as for the original parallel data.

For the final system, we train eight independent models: four left-to-right and four right-to-left, from which we chose one model checkpoint from each based on the score on newsdev2017.
The 50-best output from a left-to-right ensemble was rescored using the right-to-left models.
When scoring translation candidates, we normalise the log probabilities by translation length, adjusted according to the method described in \newcite{wu2016google}.
We optimise the length penalty (i.e., the alpha value in \newcite{wu2016google}) on newsdev2017, setting it to 0.6 for EN$\rightarrow$LV and 0.7 for LV$\rightarrow$EN.

Our EN$\rightarrow$LV models are also used in the QT21 system combination.
For a description of the combined system and results, see \newcite{Jan-Thorsten2017}.

\subsection{Russian \tofrom English}

We use the following resources from the WMT parallel data: News Commentary v12, Common Crawl, Yandex Corpus and UN Parallel Corpus V1.0.  We do not use Wiki Headlines. To increase the consistency
between English and Russian segmentation
despite the differing alphabets, we transliterate
the Russian vocabulary into Latin characters
with ISO-9 to learn the joint BPE encoding, then
transliterate the BPE merge operations back into
Cyrillic.
We apply the 
concatenation of the Cyrillic and Latin merge operations to the English and Russian side.

In order to incorporate in-domain parallel training data, we also use Edinburgh's WMT16 systems to backtranslate monolingual data. We translate the Russian (7.1M sentence) and English (20.4M sentences) News Crawl articles from 2016, which is combined with human-translated parallel data in a 1:1 mix. We used the deep transition architecture for our experiments.

For the final system, we train eight independent models: four left-to-right and four right-to-left, from which we choose one model checkpoint from each based on the score on newsdev2017.
The 50-best output from a left-to-right ensemble was rescored using the right-to-left models.
There was a preprocessing error in the RU$\rightarrow$EN backtranslation data and this is the reason that the submission result is worse than the corrected results published in this paper.

\subsection{Turkish \tofrom English}
We use all of the available parallel training data to train our TR$\leftrightarrow$EN systems. 
This consists of about 200k parallel sentences after preprocessing. 
The preprocessing is as described in section~\ref{sec:preprocess}, with the exception of the subword segmentation. 
For both directions, we do not include the modifications to subword segmentation described in section~\ref{bpe-section}; i.e. we do not disallow rare subword units in the training corpus. 
This is done because of the relatively small amount of training data for this language pair. 

In addition to the parallel training data, target-side monolingual data is incorporated into our systems. 
For both languages, we randomly select about 400k sentences from the WMT News Crawl 2016 corpus for this purpose. 
The same monolingual data is used as both back-translated and copied data (see section~\ref{sec:monolingual}), and we use a mixed training regime for all experiments. 
We create the back-translated corpus using a shallow NMT system trained on the parallel training data.

We use the stacked model architecture for all systems. 
We train eight models for each translation direction: four left-to-right and four right-to-left. 
We ensemble the left-to-right models and take the 50 best translation hypotheses; these are reranked using an ensemble of the right-to-left models.

\subsection{Biomedical Task Systems}

\subsubsection{Overview}

The systems for EN$\rightarrow$PL and EN$\rightarrow$RO are created specifically for the WMT17 biomedical task using a similar model to the systems 
created for the news task. We use all the parallel data provided in the UFAL Corpus released for this task, first removing any parallel sentences where
either side contains no ascii letters, then running the preprocessing pipeline as described in Section \ref{sec:preprocess}. For
Romanian, we apply normalisation of ``t-comma'' and ``s-comma'' characters.

For EN$\rightarrow$CS and EN$\rightarrow$DE, our systems are based on earlier work,
so are created using different data sets. For EN$\rightarrow$CS, 
our starting point is the Edinburgh WMT16 system, whereas for EN$\rightarrow$DE we use all available data from OPUS\footnote{\url{http://opus.lingfil.uu.se/}}
(gathered in May 2015) plus a small (10,000 sentence) corpus of translated Cochrane abstracts.

\begin{table*}[t]
  \small\centering
\caption{\bleu scores for translating news {\em into} English (WMT 2016 and 2017 test sets -- WMT 2017 dev set is used where there was no 2016 test)}
\label{en-*}\vspace{1em}

\begin{tabular}{@{}lcccccccccccc@{}}
  & \multicolumn{2}{c}{\bfseries CS$\to$EN}
  & \multicolumn{2}{c}{\bfseries DE$\to$EN}
  & \multicolumn{2}{c}{\bfseries LV$\to$EN}
  & \multicolumn{2}{c}{\bfseries RU$\to$EN}
  & \multicolumn{2}{c}{\bfseries TR$\to$EN}
  & \multicolumn{2}{c}{\bfseries ZH$\to$EN}\\
  \bfseries system
  & \textbf{2016} & \textbf{2017}
  & \textbf{2016} & \textbf{2017}
  & \textbf{2017d} & \textbf{2017}
  & \textbf{2016} & \textbf{2017}
  & \textbf{2016} & \textbf{2017}
  & \textbf{2017d} & \textbf{2017}\\
  \hline
  WMT-16 single system &
  30.1 & 25.9 & 
  36.2 & 31.1 & 
  --- & --- &
  26.9 & 29.6 & 
  --- & --- & 
  --- & --- \\

baseline &
31.7 & 27.5 & 
38.0 & 32.0 & 
23.5 & 16.4 & 
27.8 & 31.3 & 
20.2 & 19.7 & 
19.9 & 21.7 \\ 

+layer normalization &
32.6 & 28.2 & 
38.6 & 32.1 & 
24.4 & 17.0 & 
28.8 & 32.3 & 
19.5 & 18.8 & 
20.8 & 22.5 \\ 

+deep model &
33.2 & 28.9  & 
39.6 & 33.5 & 
24.4 & 16.6 & 
29.0 & 32.7 & 
20.6 & 20.6 & 
22.1 & 22.9 \\ 

+checkpoint ensemble &
33.8 & 29.4 & 
39.7 & 33.8 & 
25.7 & 17.7 & 
29.5 & 33.3 & 
20.6 & 21.0& 
22.5 & 23.6 \\ 

+independent ensemble &
34.6& 30.3 & 
40.7 & 34.4 & 
27.5 & 18.5 & 
29.8 & 33.6 & 
22.1 & 21.6 & 
23.4 & 25.1 \\ 

+right-to-left reranking &
35.6& 31.1 & 
41.0 & 35.1 & 
28.0 & 19.0 & 
30.5 & 34.6 & 
22.9 & 22.3 & 
24.0 & 25.7 \\ 

WMT-17 submission$^a$ &

--- & \textbf{30.9} & 
--- & \textbf{35.1} & 
--- & \textbf{19.0} & 
--- & \textbf{30.8} & 
--- & \textbf{20.1} & 
--- & \textbf{25.7}  \\ 
\hline

\multicolumn{13}{p{\linewidth - 1\tabcolsep}@{}}{\tiny
\hspace{-\tabcolsep}$^a$ In some cases training did not converge until
after the submission deadline. The contrastive/ablative results shown
were obtained with the converged systems; this line reports the \bleu
score for the system output submitted by the submission deadline.}
\end{tabular}
\end{table*}

\begin{table*}[t]
\small
\caption{\bleu scores for translating news {\em out of} English (WMT 2016 and 2017 test sets -- WMT 2017 dev set is used where there was no 2016 test)}
\label{*-en}\vspace{1em}

\noindent
\begin{tabular}{@{}lcccccccccccc@{}}
& \multicolumn{2}{c}{\bfseries EN$\to$CS}
& \multicolumn{2}{c}{\bfseries EN$\to$DE}
& \multicolumn{2}{c}{\bfseries EN$\to$LV}
& \multicolumn{2}{c}{\bfseries EN$\to$RU}
& \multicolumn{2}{c}{\bfseries EN$\to$TR}
& \multicolumn{2}{c}{\bfseries EN$\to$ZH}\\
\bfseries system & \textbf{2016} & \textbf{2017} & \textbf{2016} & \textbf{2017} & \textbf{2017d} & \textbf{2017} & \textbf{2016} & \textbf{2017} & \textbf{2016} & \textbf{2017} & \textbf{2017d} & \textbf{2017}\\
\hline

WMT16 single system &
23.7 & 19.7 & 
31.6 & 24.9 & 
--- & --- & 
24.3 & 26.7 & 
--- & --- &
--- & --- \\
baseline &
23.5 & 20.5 & 
32.2 & 26.1 & 
20.8 & 14.6 & 
25.2 & 28.0 & 
13.8 & 15.6 & 
30.5 & 31.3 \\ 
+layer normalization &
23.3 & 20.5 & 
32.5  & 26.1 & 
21.6 & 14.9 & 
25.8 & 28.7 & 
14.0 & 15.7 & 
31.6  & 32.3 \\ 
+deep model &
24.1 & 21.1  & 
 33.9 & 26.6 & 
22.3 & 15.1 & 
26.5 & 29.9 & 
14.4 & 16.2 & 
32.6 & 33.4 \\ 
+checkpoint ensemble &
24.7 & 22.0 & 
33.9 & 27.5 & 
23.4 & 16.1 & 
27.3 & 31.0 & 
15.0 & 16.7 & 
32.8  & 33.5 \\ 
+independent ensemble &
 26.4 & 22.8 & 
 35.1 & 28.3 & 
24.7 & 16.7 & 
28.2 & 31.6 & 
15.5 & 17.6 & 
35.4 & 35.8 \\ 
+right-to-left reranking &
26.7 & 22.8 & 
36.2 & 28.3 & 
25.0 & 16.9 & 
-- & -- & 
16.1 & 18.1 & 
35.7 & 36.3 \\ 
WMT-17 submission$^a$ & 
-- & \textbf{22.8} & 
--  & \textbf{28.3} & 
-- & \textbf{16.9} & 
-- & \textbf{29.8}  & 
-- & \textbf{16.5} & 
-- & \textbf{36.3} \\ 
\hline
\multicolumn{13}{p{\linewidth - 1\tabcolsep}@{}}{\tiny
\hspace{-\tabcolsep}$^a$ In some cases training did not converge until
after the submission deadline. The contrastive/ablative results shown
were obtained with the converged systems; this line reports the \bleu
score for the system output submitted by the submission deadline.}
\end{tabular}
\end{table*}

\subsubsection{Synthetic Data}
As in the news task, we seek to improve performance of the generic system by using in-domain training data, synthesising new
data when there is insufficient naturally-occurring parallel data. We first tried fine-tuning with the EMEA corpus (drug
information leaflets), but this did not give good results, probably because it is relatively small and not sufficiently
close to the domain of interest.

Turning to back-translation as a source of parallel data for fine-tuning,
we find that there is no good source of in-domain target language data. 
So, since the development and test sets are drawn from the websites of NHS 24 and Cochrane, we apply the following procedure
in order to generate in-domain synthetic data:
\begin{compactenum}
\item Crawl the NHS~24 websites (\url{www.nhsinform}, \url{www.nhs24.com}, \url{www.scot.nhs.uk}) and the Cochrane
websites (\url{www.cochrane.org} and \url{www.cochranelibrary.com}) to create English corpora of about 64k and 174k segments,
respectively.
\item Machine translate each of these crawled corpora into the 4 target languages (Czech, German, Polish and Romanian). For
all except for Polish, we used Edinburgh WMT16 system. For Polish we use a shallow Nematus system trained on OPUS.
\item Apply Moore-Lewis selection \cite{moore-lewis:2010:Short}, using the translated Cochrane and NHS 24 crawls as in-domain
data, to select from the monolingual CommonCrawl corpus \cite{Buck-commoncrawl} in each of the 4 languages. We restrict
to sentences between 10 and 80 tokens long in CommonCrawl.  We select
corpora of between 4M and 10M sentences in each of 2 domains, by 4 languages.
\item Back-translate the selected corpora to English, again using either the Edinburgh WMT16 system for the language pair in question, or a Nematus system 
trained from OPUS.
\end{compactenum}
An additional complication for Romanian is that the CommonCrawl corpus is particularly inconsistent in its use
of diacritics (this is a problem we have observed to a lesser extent in other Romanian corpora).
To fix this, we train a ``diacritiser'' for
Romanian, which is actually an NMT system mapping Romanian text with diacritics stripped, to correct Romanian text. As training
data for the diacritiser we use the Europarl, DGT and SETIMES2 corpora from OPUS. The diacritiser is applied to the CommonCrawl
text selected above.

For the Romanian system we combine the corpora selected by both Cochrane and NHS 24, and train a single adapted
system, for Polish we just use the corpus selected by the NHS 24 data, and for German and Czech we used
 the separate selected corpora to create adapted systems for each of Cochrane
and NHS 24. We show the effect of this domain adaptation in Section  \ref{sec:results}.

\subsubsection{System Details}
For all language pairs, we use the HimL tuning sets for validation, and the HimL test sets as devtest sets.\footnote{\url{http://www.himl.eu/test-sets}}
\paragraph{EN$\rightarrow$CS} We use shallow Nematus models, fine-tuned from Edinburgh's WMT16 system. There are separate fine-tuning runs for
NHS 24 and Cochrane, each using 4M sentences randomly selected from CzEng, the synthetic corpus described above, and the EMEA corpus.
The final system is an ensemble of the final 4 checkpoints.
\paragraph{EN$\rightarrow$DE} This is similar to EN$\rightarrow$CS, except that the generic training corpus consists of about 44M
sentence pairs from OPUS. For fine-tuning we use the synthetic corpus, EMEA, and 10M sentences randomly selected from the generic 
corpus. For Cochrane, we add 10k parallel sentences of abstracts from the Cochrane website.
\paragraph{EN$\rightarrow$PL} We use the UFAL corpus (39M sentence pairs) as the generic corpus, and the
synthetic data and EMEA as the in-domain data (19M sentence pairs). The final system is an ensemble of four
target left-right systems, reranked with two target left-right systems. One of each of the left-right and right-left
systems uses the fine-tuned training regime and the stacked model architecture, whereas the others use the mixed regime
and deep transition architecture.
In the reranking, we apply a heuristic to remove hypotheses consisting of many
repeated quotes, as well as a length normalisation trick \cite{wu2016google}. For the latter, we optimise alpha
on the HimL test sets, setting it to 0.6 for NHS24 and 1.2 for Cochrane.
\paragraph{EN$\rightarrow$RO} The generic data for this system is the UFAL corpus (about 62M sentence pairs) with our in-domain set
consisting of the synthetic
data created as above and EMEA (about 11M sentence pairs).
The final system is an  ensemble of three deep target left-right systems, reranked with three target right-left 
systems. The first of the left-right runs used the stacked architecture, and the fine-tuned training regime, whereas the others
used the deep transition architecture and mixed training. We again use the length normalisation trick, with alpha set to 
0.7 for NHS 24.

\section{Results}
\label{sec:results}

\begin{table*}
\begin{center}
\small
\caption{Contrastive experiments for biomedical task. Submitted system marked in bold.}
\label{bio}\vspace{1em}

\begin{tabular}{l|cccc|cccc}
\multicolumn{1}{c}{}  & \multicolumn{4}{|c|}{\bfseries EN$\to$PL} & \multicolumn{4}{c}{\bfseries EN$\to$RO} \\
\cline{2-9}
\multicolumn{1}{c}{}  & \multicolumn{2}{|c|}{devtest} & \multicolumn{2}{c|}{test} & \multicolumn{2}{c|}{devtest} & \multicolumn{2}{c}{test} \\
  \hline
\bfseries system & \bfseries Coch & \bfseries NHS24 & \bfseries Coch & \bfseries NHS24 & \bfseries Coch & \bfseries NHS24 & \bfseries Coch & \bfseries NHS24  \\
\hline
baseline &
19.8 & 24.3 & 26.2 & 18.2 &  
35.4 & 29.5  & 36.8 & 23.0 \\ 
+layer normalization &
20.3  & 24.8   & 25.5   & 20.2  & 
34.4 & 29.9  & 35.6 & 24.7 \\ 
+deep model &
20.6  & 24.5  & 25.9  & 20.2 & 
36.7  & 30.0  & 37.8 & 27.3 \\ 
+checkpoint ensemble &
21.3 & 26.3 & 28.4  & 21.3 & 
37.3 & 29.9  & 39.1  & 27.0 \\ 
+independent ensemble &
22.2  & 27.8 & 28.1  & 21.6 & 
39.1 & 32.8  & 40.5 & 28.3 \\ 
+right-to-left reranking &
22.1 & 28.2 & 28.6  & 22.5  &  
39.5 & 34.9  & 40.8 & 29.0 \\ 
\hline
WMT17 submission$^a$  & 
-- & -- &  \textbf{29.0} & \textbf{23.2} & 
-- & --  & \textbf{41.2} & \textbf{29.3} \\ 
\hline

\multicolumn{9}{p{13cm}}{\tiny
\hspace{-\tabcolsep}$^a$ For the submitted systems we show the \bleu scores provided by the organisers, which used 
a different tokenisation to the other scores in the table. The outputs are all obtained using the +right-to-left reranking system}

\end{tabular}
\end{center}
\end{table*}

The main results for the news translation task are shown in Tables \ref{en-*} and \ref{*-en}.
We report case-sensitive, detokenized \bleu, using the NIST \bleu\ scorer.\footnote{\tiny \url{https://github.com/moses-smt/mosesdecoder/blob/master/scripts/generic/mteval-v13a.pl}} For Chinese
output, we split to characters using the script supplied for WMT17 before running \bleu.

For reporting single system scores, we arbitarrily choose the first system that we trained, out of the systems using the mixed training regime.
In some cases we obtained improvements after the submission deadline, either due to longer training or preprocessing changes.
In these cases the contrastive/ablative results show the best-performing systems, but we include the \bleu of the submitted system for completeness.

For the biomedical systems, we show results on the HimL test sets (``devtest'') as well as the final released test sets in Table \ref{bio}.
The ``+right-to-left reranking'' system also introduces the tuned length normalisation.

For the language pairs for which we participated both in WMT16 and WMT17, we also show the scores of last year's systems.
We observe solid improvements over these, with improvements of 1.5--3 {\sc Bleu} for single models.
Some of these improvements are visible in our baseline systems, which indicates that they are due to differences in training data, preprocessing and hyperparameters.

We have highlighted the performance improvements of two architecture variants, layer normalization and deep models, which lead to improvements in \bleu across most language pairs.
We also show contrastive results for ensembling, comparing checkpoint ensembles to more expensive independent ensembles.
We find that checkpoint ensembles generally yield performance improvements over a single model, but that independent ensembles are consistently more effective.
Right-to-left reranking yielded an average improvement of 1 \bleu in our 2016 experiments; this year, improvements are smaller, and between 0 and 1 \bleu.
We attribute this to the stronger performance of single models.

To show the effect of our domain adaptation for the biomedical task, we display results in Table \ref{tab:domain} with and without the synthetic data. In this table, the 
``generic'' system uses just the provided parallel data, with the stacked architecture. After training to convergence on this data, we then fine-tune the best model (selected
by \bleu) using a 50-50 mixture of in-domain data (synthetic plus EMEA) and the generic parallel data. We report \bleu using the NIST scorer, for the best single model and
an ensemble of the last four checkpoints, comparing the generic and the fine-tuned systems. Note that the system shown here is a different run to the single system shown in Table \ref{bio}.

 We can see that the adaptation 
has a positive effect on \bleu on all of the EN$\rightarrow$PL test sets, however in EN$\rightarrow$RO the effect is more mixed. We note that there are improvements
on the corresponding single best models, but there seems to be a problem with the checkpoint ensemble for NHS 24. Looking more closely at the output we can see
that when the \bleu score drops, the output is around 10\% longer, due to the increased proportion of ``nonsense'' sentences. The checkpoint ensemble is
perhaps being more affected by volatility in training,
as it selects models based on iteration count, rather than heldout performance.

\begin{table*}
\begin{center}
\small
\caption{Effect of domain adaptation for biomedical task. We show the ``generic'' system (trained on parallel data only) and a ``fine-tuned'' system 
(the best generic system, fine-tuned on a mix of synthetic in-domain data, EMEA data and sampled parallel data).
 We show scores for both the single best system, and an ensemble of the last four
checkpoints.}
\label{tab:domain}\vspace{1em}

\begin{tabular}{l|cccc|cccc}
\multicolumn{1}{c}{}  & \multicolumn{4}{|c|}{\bfseries EN$\to$PL} & \multicolumn{4}{c}{\bfseries EN$\to$RO} \\
\cline{2-9}
\multicolumn{1}{c}{}  & \multicolumn{2}{|c|}{devtest} & \multicolumn{2}{c|}{test} & \multicolumn{2}{c|}{devtest} & \multicolumn{2}{c}{test} \\
  \hline
\bfseries system & \bfseries Coch & \bfseries NHS24 & \bfseries Coch & \bfseries NHS24 & \bfseries Coch & \bfseries NHS24 & \bfseries Coch & \bfseries NHS24  \\
\hline
generic (single) &
18.4 & 23.4  & 22.8  & 16.6  &  
35.6 & 30.9  & 37.6  & 26.5 \\ 
generic (ensemble 4) &
19.9 & 24.3  & 23.6  & 19.9 &  
37.5 & 33.3  & 39.2  & 27.9  \\ 
\hline
fine-tuned (single) &
 20.7 & 26.5   & 27.2  & 19.5 &  
 36.7 & 30.1  & 38.6 & 27.0  \\ 
fine-tuned (ensemble 4) &
20.8 & 26.7    & 27.4  & 20.9  &
37.9 & 32.0 & 39.9 & 26.0  \\
\hline
\end{tabular}
\end{center}
\end{table*}

An independent analysis of our EN$\rightarrow$CS and EN$\rightarrow$LV news systems' performance with regard to morphology can be found in \newcite{burlot16morpheval}.

\section{Conclusions}

This paper describes the University of Edinburgh's submissions to the WMT17 shared news translation and biomedical translation tasks.
We perform extensive ablation experiments to report the effectiveness of different architecture choices and ensembling techniques.
We report strong baselines that use both parallel and (back-translated) monolingual data, and already outperform our last year's submissions to WMT 2016.
On top of these, we find that layer normalization and deep models lead to improvements across most language pairs.
We also report performance gains from ensembling and re-ranking with right-to-left models, and find that gains have decreased slightly compared to last year's systems, despite using the more expensive strategy of ensembling independently trained models.

Among constrained submissions to the news task, our submissions are ranked tied 1st for 11 out of 12 translation directions in which we participated: EN$\to$\{CS, RU, LV, TR, ZH\}, and \{CS, DE, LV, RU, TR, ZH\}$\to$EN. In the biomedical 
task, we obtained the highest \bleu\ across all submissions, for all language/domain combinations that we submitted.

\section*{Acknowledgments}
\lettrine[image=true, lines=2, findent=1ex, nindent=0ex, loversize=.15]{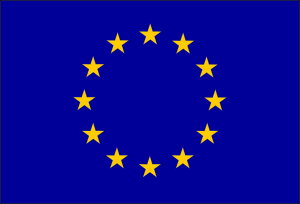}%
{T}his project has received funding from the European Union's Horizon
2020 research and innovation programme under grant agreements 645452
(QT21), 644333 (TraMOOC), 644402 (HimL), 645487 (ModernMT), and 688139
(SUMMA). 
GPU time was supported by Microsoft's donation of Azure credits to The Alan Turing Institute.  
This work was supported by The Alan Turing Institute under the EPSRC grant EP/N510129/1.

\bibliography{bibliography}
\bibliographystyle{ugHarv3plainEtal}

\end{document}